\documentclass[10pt,letterpaper]{article}

\usepackage{cogsci}

\cogscifinalcopy 

\usepackage{sidecap}
\usepackage{graphicx}
\usepackage{gensymb}
\usepackage{amsmath}
\usepackage{amsfonts}
\usepackage{lipsum}
\usepackage[breaklinks=true]{hyperref}
\usepackage{breakurl}
\usepackage{caption}
\usepackage{subcaption}
\usepackage{pslatex}
\usepackage{apacite}
\usepackage{natbib}
\usepackage{float} 

\title{Are Convolutional Neural Networks or Transformers more like human vision?}
 
\author{{\large \bf Shikhar Tuli ({stuli@princeton.edu})} \\
  Department of Electrical and Computer Engineering, Princeton University
  
  \AND {\large \bf Ishita Dasgupta ({idg@google.com})} \\
  DeepMind, New York

  \AND {\large \bf Erin Grant ({eringrant@berkeley.edu})} \\
  Department of Electrical Engineering and Computer Sciences, UC Berkeley
  
  \AND {\large \bf Thomas L. Griffiths ({tomg@princeton.edu})} \\
  Departments of Psychology and Computer Science, Princeton University
  }

\newcommand{\eg}{\textit{e.g.,}~}
\newcommand{\ie}{\textit{i.e.,}~}
\newcommand{\cf}{\textit{c.f.,}~}
\newcommand{\vs}{\textit{vs.}~}

\begin{document}

\maketitle

\begin{abstract}
Modern machine learning models for computer vision exceed humans in accuracy on specific visual recognition tasks, notably on datasets like ImageNet. However, high accuracy can be achieved in many ways. The particular decision function found by a machine learning system is determined not only by the data to which the system is exposed, but also the inductive biases of the model, which are typically harder to characterize. In this work, we follow a recent trend of in-depth behavioral analyses of neural network models that go beyond accuracy as an evaluation metric by looking at patterns of errors.  Our focus is on comparing a suite of standard Convolutional Neural Networks (CNNs) and a recently-proposed attention-based network, the Vision Transformer (ViT), which relaxes the translation-invariance constraint of CNNs and therefore represents a model with a weaker set of inductive biases. Attention-based networks have previously been shown to achieve higher accuracy than CNNs on vision tasks, and we demonstrate, using new metrics for examining error consistency with more granularity, that their errors are also more consistent with those of humans. These results have implications both for building more human-like vision models, as well as for understanding visual object recognition in humans.
\end{abstract}

\section{Introduction}
Convolutional Neural Networks (CNNs) are currently the de-facto standard for many computer vision tasks, including object detection \citep{ren2015faster}, image classification \citep{krizhevsky2012imagenet}, segmentation \citep{girshick2014rich}, facial recognition \citep{face}, and captioning \citep{chen2017sca}. The inductive bias in CNNs was inspired by the primate visual system, and their layer activations have been used to explain neural activations therein \citep{Yamins8619}.
A large amount of recent work has gone into understanding the representations and strategies learned by CNNs trained on popular datasets like ImageNet \citep{geirhos2020beyond, hermann2019exploring}. Much of this takes the form of behavioral analysis; \ie analyzing model classifications to gain insight into the underlying representations. A key finding from this work is that networks tend to classify images by texture rather than by shape \citep{baker2018deep}. On the other hand, humans preferentially use shape information for classification \citep{kucker2019reproducibility}. For example, CNNs struggle to recognize sketches that preserve the shape rather than texture, while these can easily be classified by humans. 

\begin{SCfigure}
\centering
\begin{subfigure}{.24\linewidth}
  \centering
  \includegraphics[width=\linewidth]{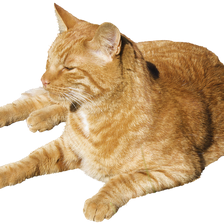}  
  \label{fig:sub-first}
\end{subfigure}
\begin{subfigure}{.24\linewidth}
  \centering
  \includegraphics[width=\linewidth]{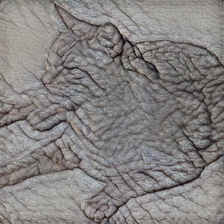} 
  \label{fig:sub-fourth}
\end{subfigure}

\vspace{-10pt}
\caption{Error-consistency stimuli \citep{geirhos2018imagenet}: (left) Original image from ImageNet, and (right) a textured transform.}
\label{fig:tex-shape_stimuli}
\end{SCfigure}

The analysis of the relationship between human vision and neural networks has been significantly improved by the availability of diagnostic datasets \citep{geirhos2018imagenet, navon1977forest}. For example, \cite{geirhos2018imagenet} present Stylized ImageNet where the texture from one class can be applied to an image from another class (preserving shape; see Figure~\ref{fig:tex-shape_stimuli}). Model performance on this dataset allows us to determine whether a model is biased toward shape or texture. Another related line of behavioral analyses instead considers error consistency on standard datasets rather than testing on specially designed datasets \citep{geirhos2020beyond}.

Recent developments in machine learning suggest, however, that convolutions maybe not be necessary for computer vision. New Transformer architectures have been successfully used for vision-based tasks \citep{vaswani2017}. These do not have the architectural inductive bias toward local spatial structure that convolutions provide. Instead, they are based entirely on flexible (learned) allocation of attention. While the success of Transformers has been most extensively demonstrated in language \citep{devlin2019bert}, their application to vision tasks have also outperformed state-of-the-art CNNs \citep{chen2020generative, pdf58:online}.

In this paper, we compare one of these attention-based models, the Vision Transformer (ViT) \citep{pdf58:online} to standard CNNs as well as humans, on a visual categorization task. We focus on understanding whether CNNs or ViTs are more ``human-like" in their classification behavior.

\section{Convolution vs. Attention}

\begin{figure}[!t]
    \centering
    \includegraphics[width=\linewidth]{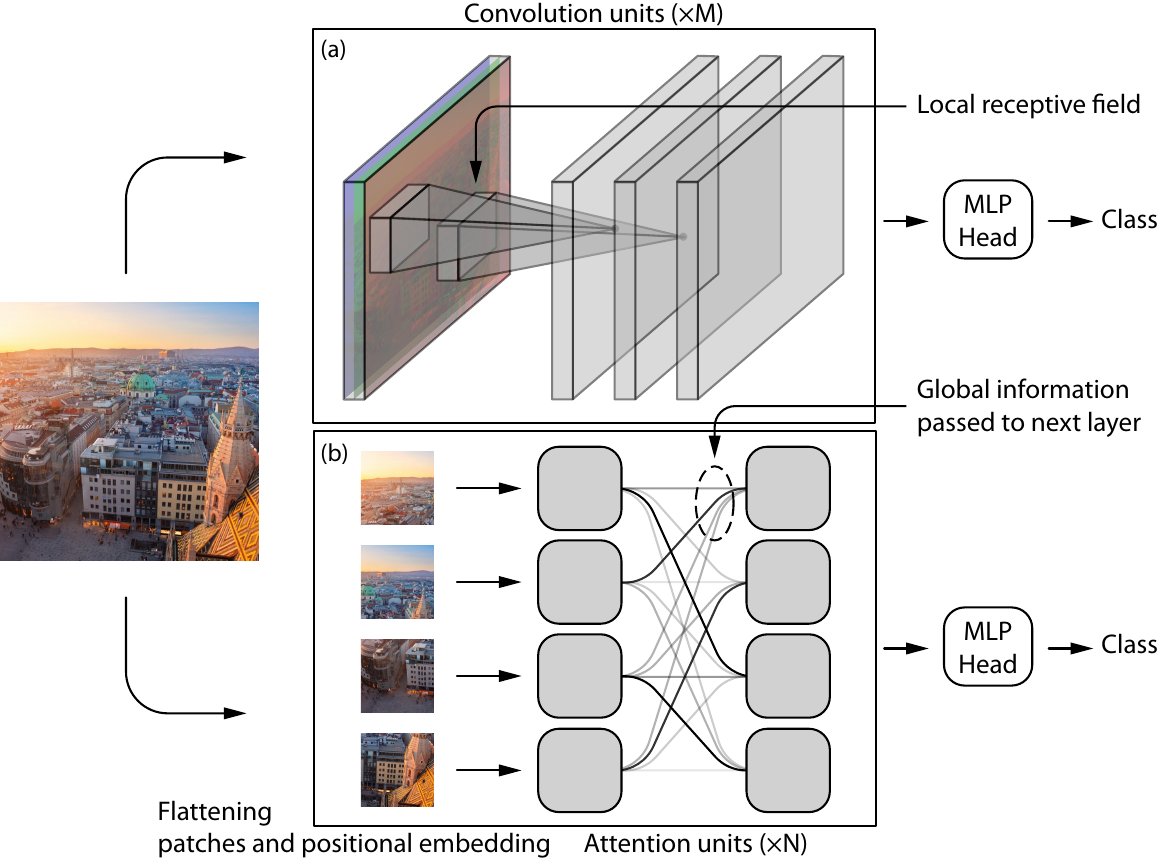}
    \caption{Bird's eye view of (a) convolutional and (b) attention-based networks.}
    \label{fig:conv_attention}
\end{figure}

Convolutional Neural Networks marked the advent of deep learning as a powerful and scalable approach \citep{krizhevsky2012imagenet} by demonstrating state of-the-art performance on large-scale image classification datasets like ImageNet. Convolutional layers convolve the input and pass its result to the next layer (see Figure~\ref{fig:conv_attention}(a)). This hard-codes a sense of translational invariance---each patch in an image is processed by the same weights. This is similar to the response of a neuron in the visual cortex to a specific stimulus \citep{lenet}. By training the weights of these convolutional filters, CNNs can learn representations of images for every specific class and have been shown to have many parallels to processing in the visual cortex \citep{Yamins8619}. These inductive biases allowed CNNs to vastly outperform fully connected networks on vision tasks. However, such local connectivity can lead to loss of global context; for example, it can encourage a bias towards classifying on the basis of texture rather than shape \citep{hermann2019exploring}. Some approaches to address this are training on augmented versions of images and incorporating top-down information \citep{cao2015look}.

Transformer models offer another approach \citep{vaswani2017}. The primary backbone of a Transformer is self-attention. This mechanism permits us to contextually up-weight the relevance of certain information. This can be used to implement local receptive fields---previous work shows that multi-head self-attention layers (like the ones we use) can perform like a convolution layer \citep{attention-conv}. However, Transformers are much more flexible and are not bound to always use convolutions. This flexibility has led to their great success in natural language processing, where one might have to attend to information at various distances away from the current word. They have recently been successful in the vision domain as well \citep{chen2020generative, pdf58:online}. In this paper, we investigate whether this added flexibility allows Transformers to give more human-like representations than CNNs.

\section{Measuring Error Consistency}

A central problem in machine learning and artificial intelligence research, as well as in cognitive science and behavioural neuroscience, is to establish whether two decision makers (be they humans or AI models) use the same strategy to solve a given task. Most comparisons across systems only consider their accuracy on the task. However, there are many ways to achieve the same average accuracy on a test set. First, two systems can differ in \emph{which} stimuli they fail to classify correctly, which is not captured by accuracy metrics. Second, while there is only one way to be right, there are many ways to be wrong---systems can also vary systematically in how they \emph{misclassify} stimuli. We consider various measures of these differences below.

\paragraph{Error overlap.}
First, we consider how to measure the similarity of two systems in terms of which stimuli they tend to misclassify. As a first pass, we can simply consider how many of the decisions down to individual trials are identical (either both correct or both incorrect). We call this the \textit{observed error overlap}. This is given by $c_{obs_{i,j}} = \frac{e_{i,j}}{n}$ where $e_{i,j}$ is how often the two systems ``agree''; \ie how often they both classify correctly or both classify incorrectly. This metric increases as the accuracies of the systems improve, since the number of overlapping correct decisions will increase. 

\paragraph{Correcting for accuracy using Cohen's $\kappa$.}
Consider a system that at each trial gets it right with probability $p_{correct}$ and gets it wrong otherwise. This amounts to taking i.i.d. samples from a binomial with parameter $p_{correct}$. Two such models will have higher observed error overlap as $p_{correct}$ (the average accuracy) increases; \ie they will have a higher \textit{error overlap expected by chance}. 
This is calculated by comparing independent binomial observers $i$ and $j$ with their accuracies as the respective probabilities: $c_{exp_{i,j}} = p_i p_j + (1-p_i)(1-p_j)$. The expected overlap can be used to normalize the observed error overlap, giving a measure of error consistency known as \textbf{Cohen's $\kappa$ }:

\begin{equation*}
    \kappa_{i, j} = \frac{c_{obs_{i, j}} - c_{exp_{i,j}}}{1 - c_{exp_{i,j}}}
\end{equation*}

Cohen's $\kappa$ has been used in previous research comparing humans and neural networks \citep{geirhos2020beyond}. However, it does not take into account what the system misclassifies an image as when making an error---it simply considers whether or not the classification was correct. It is also difficult to interpret where the similarities and differences across systems come from.

\begin{figure*}[ht]
    \centering
    \includegraphics[width=0.98\linewidth]{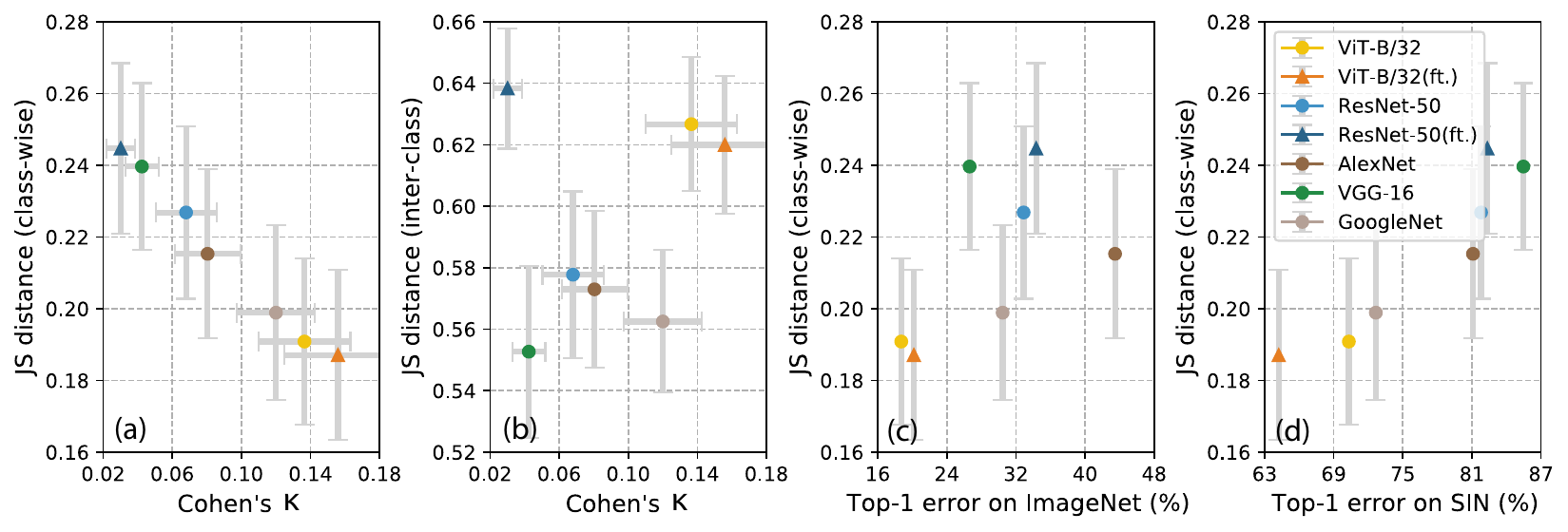}
    \caption{(a) Class-wise and (b) inter-class JS distance vs. Cohen's $\kappa$ on the Stylized ImageNet (SIN) dataset. Class-wise JS distance vs. (c) ImageNet top-1 errors and (d) SIN top-1 error.}
    \label{fig:js_vs_im&exp&cohen}
\end{figure*}

\subsection{More Granular Investigation of Misclassifications}
We can compare the decisions made by two classifiers, without loss of information, by comparing each's \emph{confusion matrix}, a table that accumulates the true \vs predicted class of each decision made by the classifier. However, this is a very high dimensional object; \eg ImageNet contains 1000 different fine-grained classes, giving a confusion matrix with $10^6$ elements. This matrix will also be very sparsely populated as most off-diagonal terms will be zero. Further, collecting adequate human data to populate the corresponding confusion matrix for human decisions is difficult.

One solution is to cluster the classes into higher-level categories, for example, by using the WordNet hierarchy \citep{miller_wordnet}; this gives 16 so-called ``entry-level'' categories, \emph{viz}. airplane, bear, bicycle, bird, boat, bottle, car, cat, chair, clock, dog, elephant, keyboard, knife, oven and truck \citep{geirhos2018imagenet}). To evaluate these ImageNet-trained models on these 16 classes, we collected the class probabilities estimated by the models and mapped these to the 16 entry-level categories by summing over the probabilities for the ImageNet classes belonging to each category, producing a 16 $\times$ 16 confusion matrix for each model.



We can use this confusion matrix to generate various metrics of comparison between two classifiers. These measures are more flexible than the Cohen's $\kappa$ metric introduced previously, since they capture information about \emph{what} misclassified elements are misclassified as \emph{what}. This opens the door to more sophisticated analyses that take into account cluster structure in the confusion matrix. For example, misclassifying a car as a truck might be a more ``human-like'' error than misclassifying it as a dog.

Concretely, we generate a probability distribution of errors over $C$ classes by computing the number of times elements from each class are misclassified and normalizing with the net number of errors made. In particular, to get a probability distribution of errors, $p \in \Delta^{C}$ (where $\Delta^C$ is the $C$-dimensional probability simplex), we normalize the error terms for every class:

\begin{equation*}
    p_i = \frac{e_i}{\sum_i^{C} e_i}, \enskip \forall \ i \in \{1, 2, \ldots, C\}~,
\end{equation*}

\noindent where $e_i$ is a count of errors defined for a given system.



We then compute the Jensen-Shannon (JS) distance between these distributions, given by

\begin{equation*}
    JS(p, q) = \sqrt{\frac{D(p \parallel m) + D(q \parallel m)}{2}}
\end{equation*} 

\noindent where $m$ is the point-wise mean of two probability distributions $p$ and $q$ (\ie $m_i = (p_i + q_i)/2$, $p$ and $q$ being the probability distributions of errors of the two systems), and $D$ is the Kullback-Leibler divergence. 
The JS distance is a symmetrized and smoothed version of the Kullback-Liebler divergence,

\begin{equation*}
    D(p \parallel q) = \sum_i p_i\text{log}\frac{p_i}{q_i}
\end{equation*} 

A lower JS distance implies classifiers with high error consistency. This measure, unlike Cohen's $\kappa$, is only concerned with similarities in which examples tend to be misclassified, and is unaffected by the overall accuracy of the classifier \citep[\cf][]{geirhos2020beyond}.
In the next section, we compute these distances to human classification behavior in convolutional and Transformer models, showing how these can yield more information than existing measures like Cohen's $\kappa$.
In what immediately follows, we define two variants of the JS distance that are less and more granular. 




\paragraph{Class-wise JS distance.} To produce an error-consistency metric that is very close to Cohen's $\kappa$, we collapse columns (predicted labels) of the confusion matrix, and compute the accumulated error for $16$ true classes as:

\begin{equation*}
    e_i = \sum_j CM_{i, j}, \enskip \forall \ j \neq i
\end{equation*}

\noindent where $CM$ is the confusion matrix for the given system in question. In this context, the class-wise JS distance compares \emph{which} classes were misclassified, for a given number of output classes (16 in this case).

\paragraph{Inter-class JS distance.}
We can also use the confusion matrix to compute more fine-grained measures. In particular, we can directly compute the distances between the full distribution of errors giving a 240-dimensional inter-class error distribution (\ie $p \in \Delta^{240}$ corresponding to the off-diagonal entries of the $16 \times 16$ confusion matrix) by taking the error counts to be the off-diagonal elements of the confusion matrix:

\begin{equation*}
    e_{ij} = CM_{i, j}, \enskip \forall \ j \neq i
\end{equation*}

\noindent In this context, the inter-class JS distance compares \emph{what} classes were misclassified as \emph{what}.

 An interesting finding is that, instead of a strong correlation shown by class-wise JS in Figure~\ref{fig:js_vs_im&exp&cohen}(a), Figure~\ref{fig:js_vs_im&exp&cohen}(b) suggests that there is no correlation of inter-class JS distance with Cohen's $\kappa$ implying that this metric gives insight beyond Cohen's $\kappa$ in measuring error-consistency with humans. 

\subsection{Methods}

We analyze error consistency for different algorithms---the most popular CNN, \ie ResNet \citep{kolesnikov2019big}, and the recently proposed attention-based Vision Transformer (ViT) \citep{pdf58:online}. The ViT and ResNet models used were pre-trained on ImageNet-21K (also known as the ``Full ImageNet, Fall 2011 release'') and ILSVRC-2012 datasets \citep{ILSVRC15}. The ViT models used include ViT-B/16, ViT-B/32, ViT-L/16 and ViT-L/32 and the ResNet model used is BiT-M-R50x1\footnote{Trained models available at \url{https://console.cloud.google.com/storage/browser/vit_models/} and \url{https://console.cloud.google.com/storage/browser/bit_models/}}.
We test these on a specially designed diagnostic dataset, the Stylized ImageNet dataset where cue-conflict between texture and shape are generated by texture-based style transfer \citep{geirhos2018imagenet}\footnote{Data available at \url{https://github.com/rgeirhos/texture-vs-shape/tree/master/stimuli}}. All results are reported with 95\% confidence intervals on cross-validated test error.





\begin{figure}[!t]
    \centering
    \includegraphics[width=\linewidth]{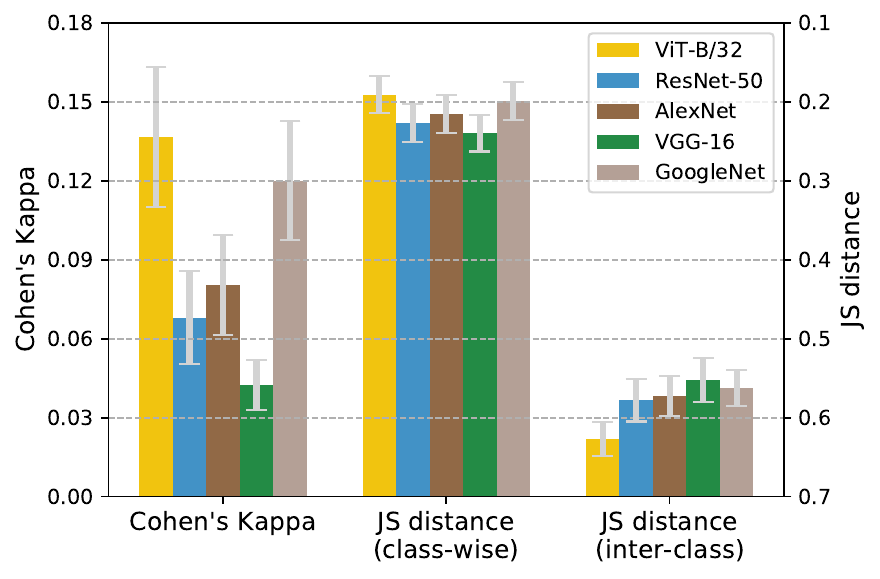}
    \caption{Error consistency results on SIN dataset.}
    \label{fig:all_EC_sin}
\end{figure}

\subsection{Results}
Firstly, we see that the class-wise JS distance is closely (inversely) correlated with Cohen's $\kappa$ when evaluated on a range of models on the Stylized ImageNet (SIN) dataset (Figure~\ref{fig:js_vs_im&exp&cohen}(a)). We also see that this measure does not have significant correlation with accuracy on the training task (Figure~\ref{fig:js_vs_im&exp&cohen}(c)), reinforcing that error consistency on diagnostic datasets like SIN can be independent of accuracy on the training task. However, when comparing this measure with the accuracy on SIN (Figure~\ref{fig:js_vs_im&exp&cohen}(d)), we can see that a lower JS distance with human errors implies a higher accuracy on SIN; \ie a higher shape bias (see the ``Shape Bias'' section below).

Figure~\ref{fig:all_EC_sin} presents the comparison of different error consistency metrics for the considered models on the SIN dataset. We first compare Cohen's $\kappa$ across ResNet and ViT. In this first comparison of the error consistency between Transformers and CNNs, we find that ViT is more consistent with humans than ResNet. We then compare our new JS distance measures. We first consider the class-wise distance. Higher Cohen $\kappa$ and lower JS distance each indicate greater error consistency. We plot the JS distance in decreasing magnitude to visually highlight similarities in the pattern with Cohen's $\kappa$. We find that the pattern repeated for Cohen's $\kappa$ is replicated with the class-wise JS distances; \ie that ViT is more human-like than CNNs. Finally, we consider the distance between the full joint distributions or the inter-class JS distance for 240 error types. We note the surprising finding that the inter-class JS distance for ViT is higher than for ResNet. 


\begin{figure}[!t]
    \centering
    \includegraphics[width=\linewidth]{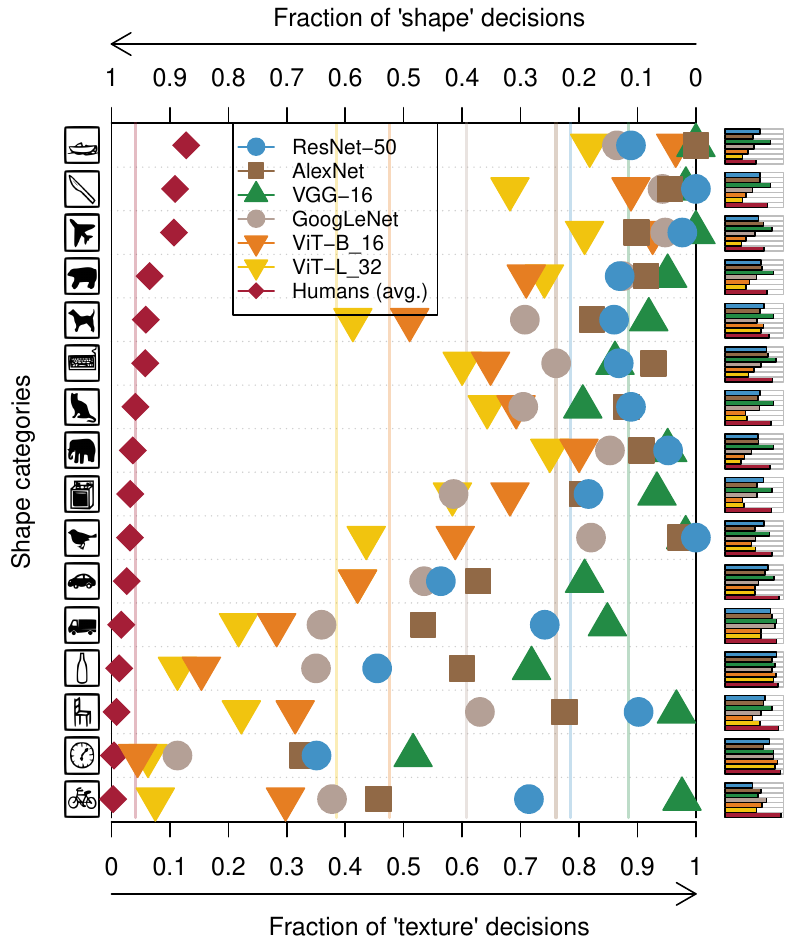}
    \caption{Shape bias for different networks for the SIN dataset \citep{geirhos2018imagenet}. Vertical lines indicate averages.}
    \label{fig:shape_bias_sin}
\end{figure}

\begin{figure}[!t]
    \centering
    \includegraphics[width=\linewidth]{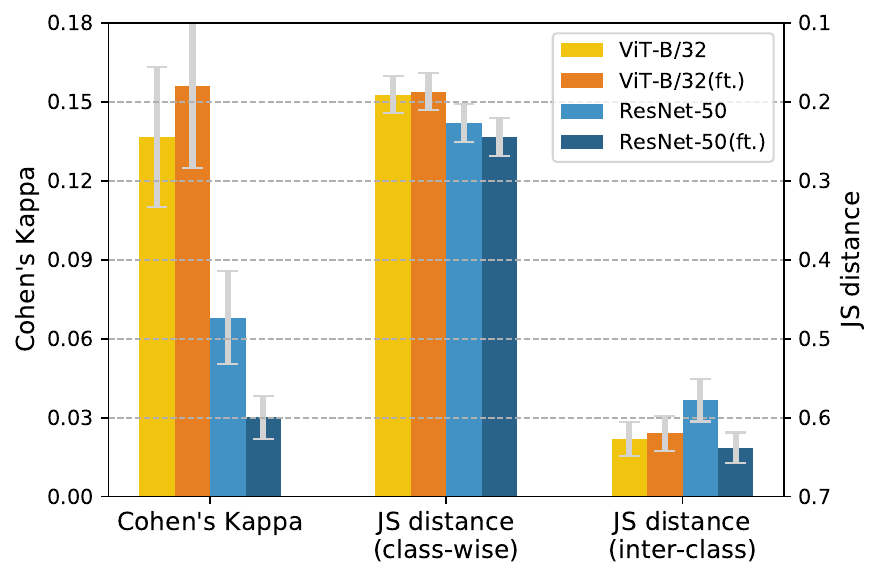}
    \caption{Error consistency results for SIN dataset before and after fine-tuning.}
    \label{fig:tex-shape_sin_ft}
\end{figure}

\section{Shape Bias}
CNNs have been shown to have a stronger dependence on texture rather than shape when categorizing visual objects \citep{baker2018deep}, while humans tend to have the opposite preference. Here, we examine how ViTs and CNNs compare on this \textit{shape bias}. The shape bias has been defined as the percentage of the time the model correctly predicts the shape for trials on which either shape or texture prediction is correct \citep{geirhos2020beyond}. We tested this by evaluating performance on the SIN dataset. This dataset contains images where the shape and texture of the object in each image conflict. With this dataset, we can test if a system classifies on the basis of shape or texture.

We use the same models as in the previous section, trained in the same way. We then analyze their test performance on the SIN dataset. We first collect all the trials in which the classifier label matches either the true texture or the true shape of the object. We then check what fraction of these got the shape right \vs the texture right. The results for this test are presented in Figure~\ref{fig:shape_bias_sin}. Small bar plots on the right indicate accuracy (answer corresponds to either correct texture or shape category). We see that ViT has a higher shape bias than traditional CNNs. This goes part of the way toward explaining the higher error consistency on the class-wise measures from the previous section with humans who primarily categorize objects by shape rather than texture. However, it seems at odds with the finding that when considering the full error distribution, ResNet is more human-like than ViT. Trying to understand this, we note that the shape bias analysis only considers the cases where either the shape or texture is correctly predicted. It only includes misclassifications that matched the image's true texture. It therefore doesn't contain most of the misclassifications that the full error distribution reflects. Hence, in the full error distribution, ResNet indeed outperforms ViT and this could only be revealed by the inter-class JS distance.

\section{Fine-tuning with Augmented Data}
\label{sec:ft}

\begin{figure}[!t]
    \centering
    \includegraphics[width=\linewidth]{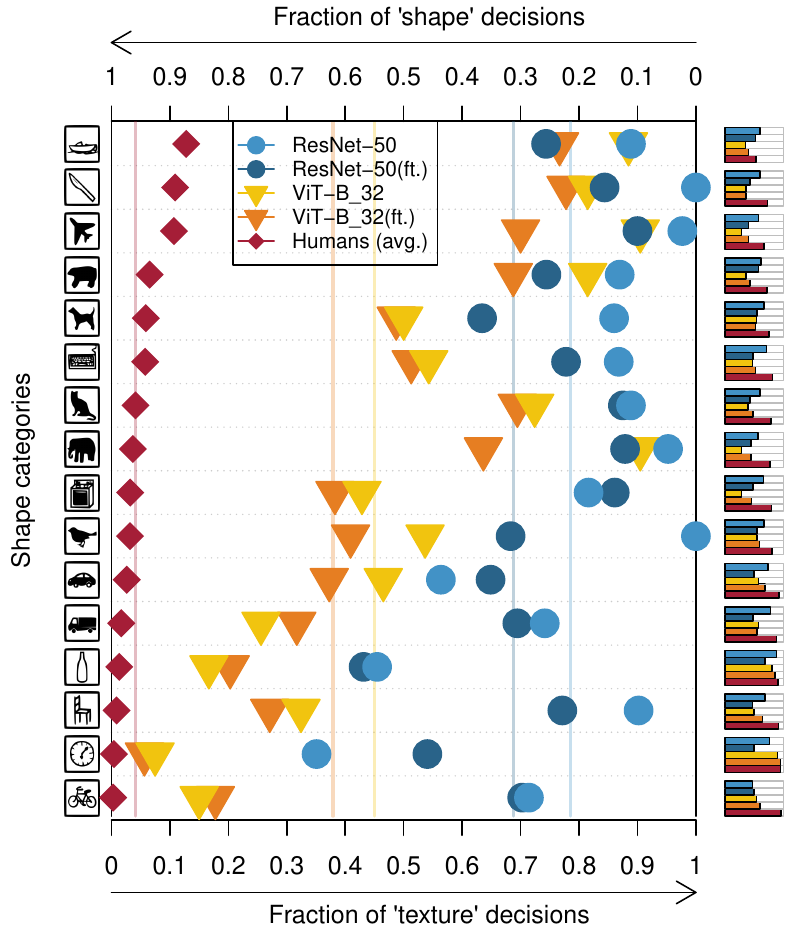}
    \caption{Shape bias for ResNet and ViT before and after fine-tuning. Vertical lines indicate averages.}
    \label{fig:shape_bias_sin_ft}
\end{figure}

We have so far only considered the performance of models that have been trained on the same fixed computer visions datasets. However, the structure of the data observed can significantly alter the representations learned. \citet{brendel2019approximating} show that ImageNet can be largely be solved using only local information, indicating that a texture bias is an ecologically rational heuristic when solving this dataset. \citet{geirhos2018imagenet} show that training on a dataset where textures are forced to be uninformative by changing the texture of ImageNet object to those of randomly chosen paintings leads to a shape bias almost as high as in humans. \citet{hermann2019exploring} show that simpler data augmentations also play a significant role in learning a shape bias. In particular, naturalistic data augmentation involving color distortion, noise, and blur substantially decreases texture bias, whereas random-crop augmentation increases texture bias in ImageNet-trained CNNs \citep{hermann2019exploring}. These findings highlight the crucial role of training data in the representations learned. In this section, we examine how data augmentations affect the representations learned by these systems, in particular, how this fine-tuning affects their similarity to human behavior.



\subsection{Methods}

Among the attention- and convolution-based models as identified previously, we train the smallest ViT model (ViT-B/32) and the smallest ResNet (BiT-M-R50x1).
Further works could look at the effect of the number of trainable parameters while training models of both architectures.

We use augmentations presented in \citet{simclr} and \citet{hermann2019exploring}: rotation ($\pm 90\degree$, $180\degree$ randomly), random cutout (rectangles of size $2\times2$ px to half the image width), Sobel filtering, Gaussian blur (kernel size = $3\times3$ px), color distortion (color jitter with probability $80\%$ and color drop with probability $20\%$) and Gaussian noise (standard deviation of 0.196 for normalized image). 
These augmentations are applied to the ImageNet dataset, and are then used to fine-tune the models. For ViT, we used cosine step decay and trained for 100 epochs with an initial learning rate of 0.3. For ResNet, we used linear step decay and trained  for 5000 epochs with an initial learning rate of 0.03. An appropriate training recipe was chosen based on the network architectures and their hyper-parameters used for the pre-training \citep{pdf58:online, kolesnikov2019big}. 
Further details about hyper-parameter tuning, along with the source code for training and the experiments implemented, can be found at: \url{https://github.com/shikhartuli/cnn_txf_bias}.



\subsection{Error Consistency}

We repeat the error consistency analyses above with these fine-tuned models. These results are reported in Figure \ref{fig:tex-shape_sin_ft}.  We find that fine-tuning made ResNet less human-like in terms of error consistency (significant differences in Cohen's $\kappa$ and the inter-class JS distance, a non-significant trend in the class-wise JS distance). This is surprising, since these augmentations have been found to increase shape bias (see next section) and one would expect that increased shape bias would mean greater error consistency with humans. On the other hand, we find that ViT does not significantly change in its error consistency with fine-tuning, and in fact trends (not statistically significantly) towards in the opposite direction than ResNet, in particular, towards improved error consistency.

\begin{figure}
    \centering
    \includegraphics[width=\linewidth]{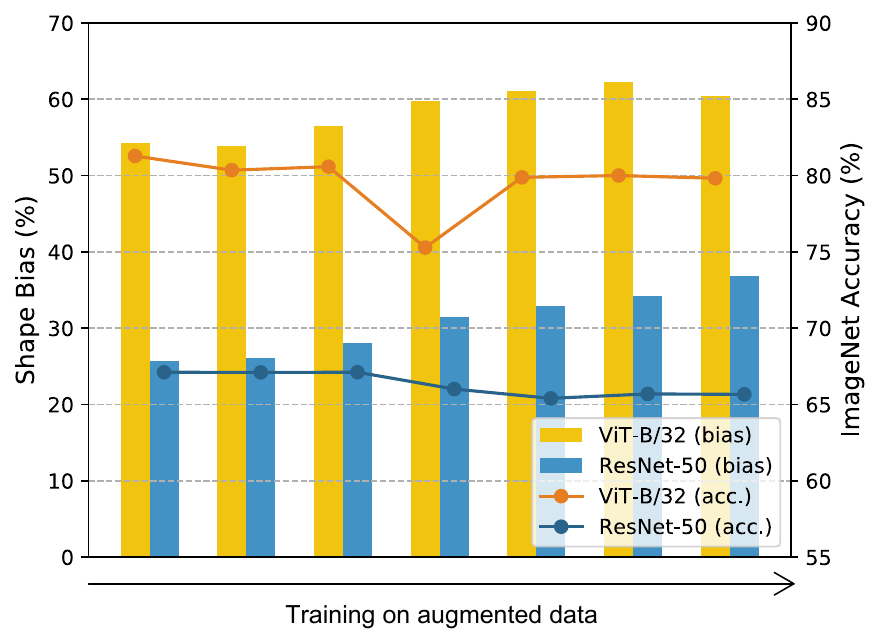}
    \caption{Shape bias and ImageNet accuracy of ViT and ResNet with fine-tuning over augmented data.}
    \label{fig:shape-bias_acc}
\end{figure}

\subsection{Shape Bias}

We repeat the shape-bias analyses above with these fine-tuned models, results in Figure \ref{fig:shape_bias_sin_ft}. Consistent with previous findings in \citet{hermann2019exploring}, we see that ResNet increases its shape bias after fine-tuning. We find that ViT also increases its shape bias after fine-tuning.

\subsection{Effect on Accuracy}
Fine-tuning changes the representations used and we have studied how this affects ML systems' similarity to human classification behavior. However, it remains to be seen how this fine-tuning affects accuracy on the original training task. We analyze this in Figure~\ref{fig:shape-bias_acc}. We see that training on augmented data increases shape bias and decreases ImageNet accuracy slightly, as is corroborated by previous works \citep{hermann2019exploring}. The decrease in accuracy for ResNet is more significant than for ViT.

\section{Conclusion}

In this work, we explore the extent to which different vision models correlate with human vision from an error-consistency point-of-view. We see that recently proposed Transformer networks not only outperform CNNs on accuracy for image classification tasks, but also have higher shape bias and are largely more consistent with human errors. We explore this consistency with new metrics that go beyond the previously proposed Cohen's $\kappa$. Further, we fine-tuned two models---a Transformer and a traditional Convolutional Neural Network (CNN)---on augmented datasets to find that this increases shape bias in both CNNs and Transformers. We observe that Transformers maintain their accuracy while also gaining equivalently in their shape bias when compared to CNNs. This could possibly be explained by the nature of attention models that permits focus on the part of the image that is important for the given task and neglect the otherwise noisy background to make predictions.

Many more tests can still be performed on Transformer models. For example, ViTs could be compared with iGPT \citep{chen2020generative} to see how the architecture within this family could affect shape and texture biases. This could help us formulate architectural ``features'' that help in modelling better brain-like networks. Additionally, the JS metric we introduce can be used to analyze the kinds of errors made in many other ways. For example, we can also gauge ``concept-level'' similarity between model misclassifications (like dogs for cats and not trucks). This could also help probe into the premise that humans could be using not just shape/texture but also ``concepts'' for classification \citep{conceptnet}. This would not be possible with a scalar metric like Cohen's $\kappa$. Further, by applying these human error-consistency metrics as part of the training loss, we could get a model nearer to ``human-like strategy''. This could also contribute towards simplifying or regularizing these models, reducing the computational cost of training.



\clearpage
\section{Acknowledgements}

This work was supported by the Defense Advanced Research Projects Agency (DARPA) under the Lifelong Learning Machines (L2M) program via grant number HR001117S0016 and by the National Science Foundation (NSF) under grant number 1718550.

\bibliographystyle{apacite}
 \def\thebibliography#1{\section*{References}
 \fontsize{10}{10}\selectfont
  \list
  {[\arabic{enumi}]}{\leftmargin \parindent
 	 \itemindent -\parindent
 	 \parsep 0.2ex plus 1pt minus 1pt
 	 \usecounter{enumi}}
 	 \def\newbrick{\hskip .11em plus .33em minus .07em}
 	 \sloppy\clubpenalty4000\widowpenalty4000
 	 \sfcode`\.=1000\relax}

 \setlength{\bibleftmargin}{.125in}
 \setlength{\bibindent}{-\bibleftmargin}

\bibliography{biblio}

\end{document}